\documentclass[letterpaper]{article} % DO NOT CHANGE THIS
\usepackage[]{aaai23}  % DO NOT CHANGE THIS
\usepackage{times}  % DO NOT CHANGE THIS
\usepackage{helvet}  % DO NOT CHANGE THIS
\usepackage{courier}  % DO NOT CHANGE THIS
\usepackage[hyphens]{url}  % DO NOT CHANGE THIS
\usepackage{graphicx} % DO NOT CHANGE THIS
\usepackage{natbib}  % DO NOT CHANGE THIS AND DO NOT ADD ANY OPTIONS TO IT
\usepackage{caption} % DO NOT CHANGE THIS AND DO NOT ADD ANY OPTIONS TO IT
\usepackage{algorithm}
\usepackage{algorithmic}

\usepackage{newfloat}
\usepackage{listings}
\DeclareCaptionStyle{ruled}{labelfont=normalfont,labelsep=colon,strut=off} % DO NOT CHANGE THIS
\floatstyle{ruled}
\newfloat{listing}{tb}{lst}{}
\floatname{listing}{Listing}
\usepackage{amsmath}
\usepackage{tabto}
\usepackage{subfigure}
\usepackage{multirow}
\usepackage{microtype}
\usepackage{booktabs}
\usepackage{enumitem}
\usepackage[utf8]{inputenc}
\usepackage{listings}
\usepackage{caption}
\usepackage{dsfont}
\usepackage{graphicx}
\usepackage{subfig}
\usepackage{amsthm}
\usepackage{afterpage}
\usepackage{xspace}
\usepackage{adjustbox}
\usepackage{booktabs, tabularx}
\usepackage{makecell}
\usepackage{nicematrix,tikz}
\usepackage{color}
\usepackage{babel}
\usepackage{nameref}

\usepackage{amsmath,amsfonts,bm}

\def\eqref#1{equation~\ref{#1}}
\def\1{\bm{1}}

\def\vx{{\bm{x}}}

\DeclareMathAlphabet{\mathsfit}{\encodingdefault}{\sfdefault}{m}{sl}
\SetMathAlphabet{\mathsfit}{bold}{\encodingdefault}{\sfdefault}{bx}{n}

\def\gA{{\mathcal{A}}}
\def\gE{{\mathcal{E}}}
\def\gM{{\mathcal{M}}}
\def\gV{{\mathcal{V}}}
\newtheorem{definition}{Definition}

\newtheorem{problem}{Problem}

\title{Heterogeneous Graph Masked Autoencoders}

\author {
    Yijun Tian\textsuperscript{\rm 1},
    Kaiwen Dong\textsuperscript{\rm 1}, 
    Chunhui Zhang\textsuperscript{\rm 2}, 
    Chuxu Zhang\textsuperscript{\rm 2},
    Nitesh V. Chawla\textsuperscript{\rm 1}
}
\affiliations {
    \textsuperscript{\rm 1} Department of Computer Science, University of Notre Dame, USA\\
    \textsuperscript{\rm 2} Department of Computer Science, Brandeis University, USA\\
    \{yijun.tian, kdong2\}@nd.edu, \{chunhuizhang, chuxuzhang\}@brandeis.edu, nchawla@nd.edu
}

\begin{document}

\maketitle

\begin{abstract}

Generative self-supervised learning (SSL), especially masked autoencoders, has become one of the most exciting learning paradigms and has shown great potential in handling graph data. However, real-world graphs are always heterogeneous, which poses three critical challenges that existing methods ignore: 
1) \textit{how to capture complex graph structure?}
2) \textit{how to incorporate various node attributes?}
and 3) \textit{how to encode different node positions?}
In light of this, we study the problem of generative SSL on heterogeneous graphs and propose HGMAE, a novel heterogeneous graph masked autoencoder model to address these challenges. HGMAE captures comprehensive graph information via two innovative masking techniques and three unique training strategies. In particular, we first develop metapath masking and adaptive attribute masking with dynamic mask rate to enable effective and stable learning on heterogeneous graphs. We then design several training strategies including metapath-based edge reconstruction to adopt complex structural information, target attribute restoration to incorporate various node attributes, and positional feature prediction to encode node positional information. Extensive experiments demonstrate that HGMAE outperforms both contrastive and generative state-of-the-art baselines on several tasks across multiple datasets. Codes are available at \url{https://github.com/meettyj/HGMAE}.

\end{abstract}

% $$$$$$$$$$$$$$$$$$$$$$$$$$$$$$$$$$$$$$$$$$$$$$$$$$$$$$$$$$
\section{Introduction}
\label{sec:intro}

Heterogeneous graphs are ubiquitous in the real world with their ability to model heterogeneous relationships among different types of nodes, such as academic graphs \cite{han}, social graphs \cite{social_graph_2}, biomedical graphs \cite{biomedical_graph_2}, and food graphs \cite{recipe2vec}. Accordingly, many heterogeneous graph neural networks (HGNNs) are proposed \cite{hgt, rn2vec, hetgnn} to capture the complex structure and rich semantics in heterogeneous graphs. Generally, HGNNs achieve great success in handling graph heterogeneity and have been applied in various real-world applications such as recommendation systems \cite{reciperec} and healthcare systems \cite{health_care_sys}. However, most HGNNs adhere to the supervised or semi-supervised learning paradigms \cite{heco}, in which the learning process is guided by labeled data. Nevertheless, the strict requirement for labels is impractical because acquiring labels in real applications is always challenging and expensive \cite{label_costly}. Therefore, self-supervised learning (SSL), which attempts to extract information from the data itself, becomes a promising solution when no or few explicit labels are provided.

In the context of SSL on graphs, contrastive SSL methods are the dominant approaches \cite{gcc, graphCL}. The success of contrastive methods is largely built upon manually selected high-quality data augmentations and complex optimization algorithms \cite{graphmae}. However, most of the augmentations on graphs are based on heuristics, with performance varying significantly across different graphs and tasks \cite{joao, data_aug_survey, graphCL}. On the other hand, complicated optimization algorithms such as momentum and exponential moving average for parameter update are always required due to the computational constraint and the need for stable training \cite{gcc, bgrl}. In addition, negative sampling, as a necessity for the majority of contrastive objectives, often involves laborious designs and arduous constructions from graphs. Therefore, there is no assurance that contrastive methods will achieve satisfactory results.

\begin{figure}[t]
	\centering
	\includegraphics[width=0.8\columnwidth]{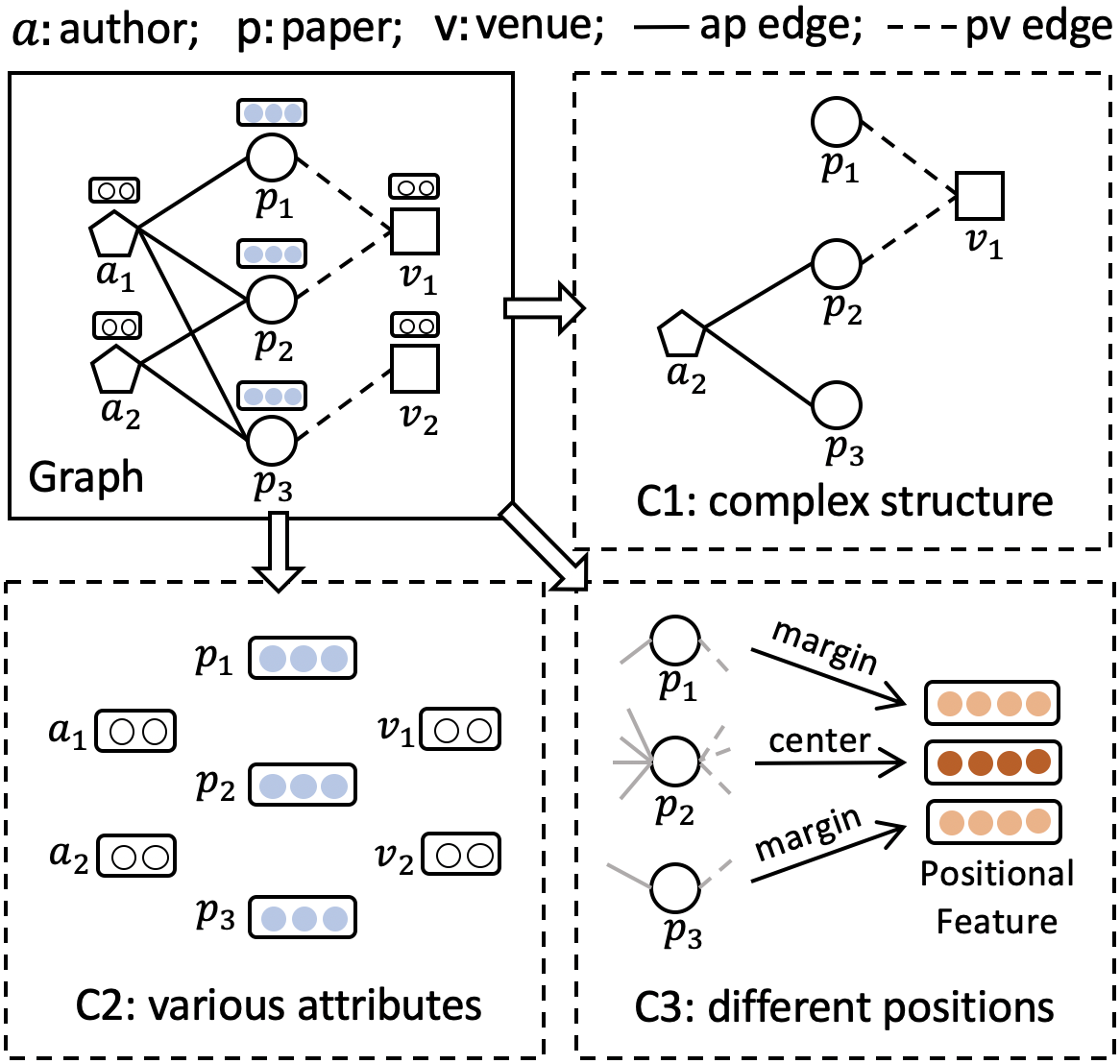}
	\caption{
	The illustration of three challenges in real-world graphs: C1 - complex structure with heterogeneous types of edges; C2 - various attributes that each node type associates; C3 - different positions that every node holds in the graph.
	}
	\label{fig:challenges}
\end{figure}

Generative SSL naturally avoids the aforementioned issues of contrastive methods, by focusing on directly reconstructing the input graph data without considering complicated augmentations and negative sampling \cite{gen_or_con, ssl_on_graph}. Recently, masked autoencoders demonstrate strong learning capability in computer vision \cite{mae} and natural language processing \cite{bert} by removing a large proportion of the input data and using the removed content to guide the training. Inspired by this, GraphMAE \cite{graphmae} proposes to reconstruct node features with masking on graphs. Although GraphMAE shows excellent performance in certain graph learning tasks, it ignores the following challenges faced for real-world graphs and limits its applicability to real applications.
\begin{itemize}
    \item \textbf{C1}: Real-world graphs are usually heterogeneous, with multiple types of nodes and edges, which naturally implies their complex structure. To learn effective node embeddings that fully consider the semantics involved in this complex structure, simply reconstructing the masked features is far from sufficient. Thus the challenge 1 is: how to capture the complex graph structure that contains informative semantics, as indicated by C1 in Figure \ref{fig:challenges}.
    
    \item \textbf{C2}:
    Different types of nodes are associated with various types of attributes. It is inappropriate to fixate the attributes of every node type and apply the same masking strategy on top of them. Thus the challenge 2 is: how to determine which node type to mask and how to design a suitable masking strategy, as indicated by C2 in Figure \ref{fig:challenges}.
    
    \item \textbf{C3}: Different nodes can carry different structural roles and be influenced by other nodes with respect to their positions in the graph. In addition to graph structure and attributes, it is important to encode the node positional information from the graph. Thus the challenge 3 is: how to incorporate different node positions and design learning objectives on top of it, as indicated by C3 in Figure \ref{fig:challenges}.
\end{itemize}

% -------------- challenges ----------------

To address these challenges, we propose HGMAE, a novel heterogeneous graph masked autoencoder for generative SSL on graphs. HGMAE contains two innovative masking techniques and three unique training strategies. Specifically, we first develop metapath masking and adaptive attribute masking to enable the encoding of complex information in heterogeneous graphs. We also employ a dynamic mask rate in adaptive attribute masking to ensure more effective and stable learning. In addition, we validate the effectiveness of leaving unchanged and replacing for masking in the graph domain, and reach a conclusion that differs from GraphMAE. Then, we design a metapath-based edge reconstruction strategy to capture the high-order relations through metapaths and encode complex graph structural information. Next, we introduce a target attribute restoration strategy to reconstruct the attributes of target nodes. The model is encouraged to fully grasp the knowledge in attributes and maintain the learning capability across different masking settings. After that, we develop a positional feature prediction strategy to fully capture the node positional information. Finally, we design a novel combined objective function of different strategies to optimize the model. To summarize, our major contributions in this paper are as follows:
\begin{itemize}
    \item To our best knowledge, this is the first attempt to study generative self-supervised learning on heterogeneous graphs. We identify and address three critical challenges pertaining to this problem.
    \item We develop two novel masking techniques including metapath masking and adaptive attribute masking. In addition, we introduce a dynamic mask rate and verify the effectiveness of leaving unchanged and replacing in the graph domain, which contradicts to previous findings.
    \item We propose HGMAE, a novel heterogeneous graph masked autoencoder model to address the challenges. HGMAE can capture comprehensive graph information including complex structure, various attributes and different node positions via several training strategies.
    \item Extensive experiments demonstrate the superiority of HGMAE compared to contrastive and generative state-of-the-art baselines on several tasks across multiple datasets.

\end{itemize}

% $$$$$$$$$$$$$$$$$$$$$$$$$$$$$$$$$$$$$$$$$$$$$$$$$$$$$$$$$$
\section{Related Work}
This work is closely related to heterogeneous graph neural networks and self-supervised learning on graphs.

\noindent
\textbf{Heterogeneous Graph Neural Networks.}
In the past few years, many heterogeneous graph neural networks have been developed to learn node representations in heterogeneous graphs \cite{hetgnn_survey_1, hetgnn_survey_2}. For example, HAN \cite{han} proposes hierarchical attention to encode node-level and semantic-level structures. MAGNN \cite{magnn} leverages the node attributes and intermediate nodes of metapaths to encapsulate rich semantic information of heterogeneous graphs. HGT \cite{hgt} introduces a transformer-based architecture to handle web-scale heterogeneous graphs. HetGNN \cite{hetgnn} proposes to capture both structure and content information in heterogeneous graphs. However, the aforementioned methods can only learn when node labels are provided and are unable to extract supervised signals from the data itself to learn general node embeddings. Therefore, how to exploit the complex information involved in heterogeneous graphs as self-supervised guidance remains a question.

\noindent
\textbf{Self-supervised Learning on Graphs.}
Self-supervised methods on graphs can be naturally divided into contrastive and generative approaches \cite{ssl_graph_survey_1, ssl_graph_survey_2}. Contrastive graph learning encourages alignment between different augmentations or distributions. For example, GCC \cite{gcc} focuses on aligning different local structures of two sampled subgraphs. GraphCL \cite{graphCL} introduces the alignment between different graph augmentations. HeCo \cite{heco} applies two network structures to align both local schema and global metapath information. However, the performance of these contrastive graph learning models highly depends on the manually selected data augmentations, which vary from graph to graph. On the other hand, generative graph learning aims to recover missing parts of the input data. In previous research, the performance of generative methods falls behind the contrastive methods \cite{gae, gala, ep}. Recently, GraphMAE \cite{graphmae} leverages masked autoencoder to reconstruct features, achieving outstanding performance. However, GraphMAE disregards several challenges in real-world data such as the complex structure, various attributes and different node positions, which we address in this paper.

% ------------- figure: framework -----------------
\begin{figure*}[t]
\begin{center}
\includegraphics[width=\textwidth]{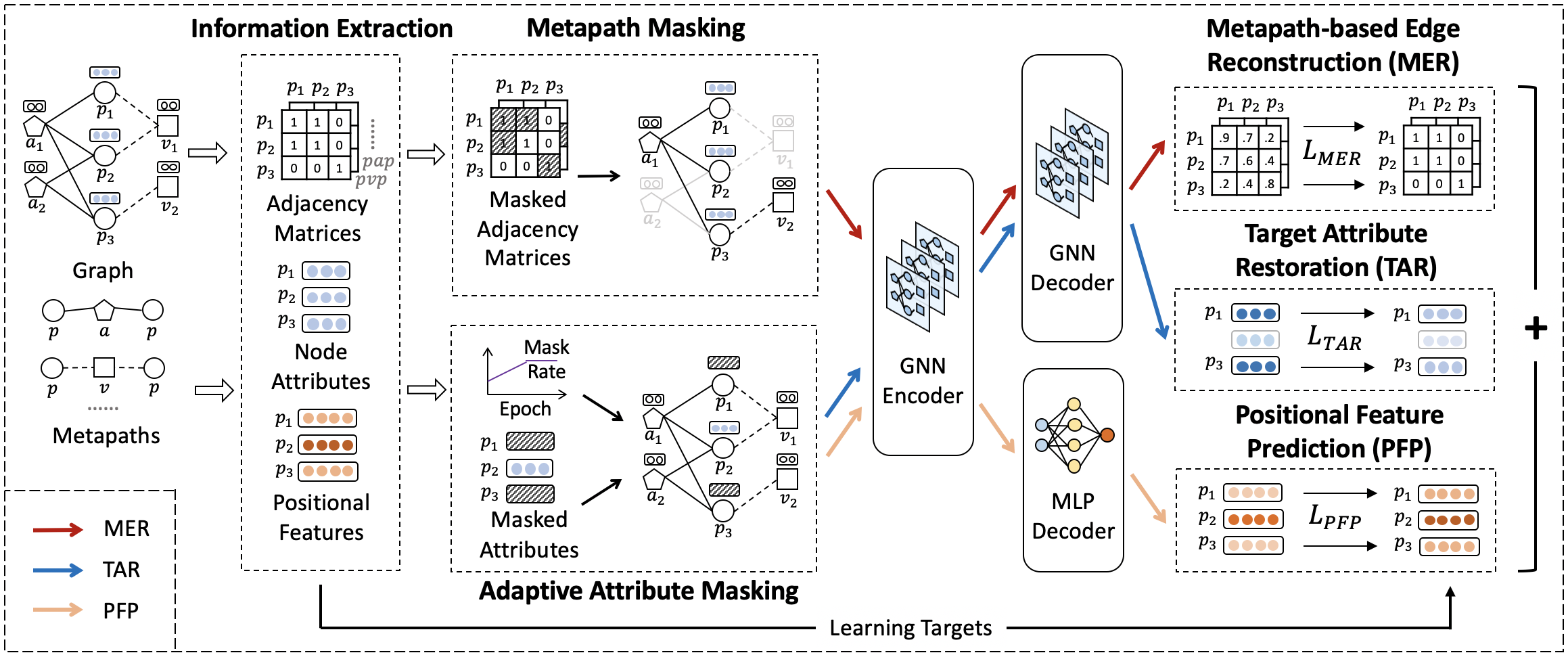}
\end{center}
\caption{
The overall framework of HGMAE: we first extract the metapath-based adjacency matrices, node attributes and positional features from the graph. We then design two masking techniques (i.e., metapath masking and adaptive attribute masking) to mask the inputs. Later, we feed the masked inputs into the encoder and decoders sequentially, and optimize them via three training strategies (i.e., metapath-based edge reconstruction, target attribute restoration, and positional feature prediction). The proposed strategies enable the model to capture comprehensive graph information and address identified challenges. 
}
\label{fig:pipeline}
\end{figure*}

% $$$$$$$$$$$$$$$$$$$$$$$$$$$$$$$$$$$$$$$$$$$$$$$$$$$$$$$$$$
\section{Problem Definition}
In this section, we describe the concept of heterogeneous graph and formally define the problem of generative self-supervised learning on heterogeneous graphs.

\begin{definition}
{\bf Heterogeneous Graph.} 
A heterogeneous graph is defined as a graph $G = (\gV, \gA, T_\gV, T_\gE, X, \Phi)$ with multiple types of nodes $\gV$ and adjacency matrix $\gA$, where $T_\gV$ represent the set of node types and $T_\gE$ is the set of edge types in $\gA$. $X$ denotes the attributes and $\Phi$ represents the metapaths. A metapath $\phi \in \Phi$ is a path that connects different types of nodes with distinct types of edges, i.e., $T_{\gV_1}\stackrel{T_{\gE_1}}{\longrightarrow}T_{\gV_2}\stackrel{T_{\gE_2}}{\longrightarrow}\dots\stackrel{T_{\gE_l}}{\longrightarrow}T_{\gV_{(l+1)}}$, where $l$ is path length.

\end{definition}

\begin{problem}
{\bf Generative Self-supervised Learning on Heterogeneous Graphs.}
Given a heterogeneous graph $G = (\gV, \gA, T_\gV, T_\gE, X, \Phi)$, the task is to first design an encoder $f_E$ that learns node embeddings $H=f_E(G)$, which encodes complex heterogeneous graph information such as structure, attributes, and node positions. Then, a decoder $f_D$ is designed to reconstruct the input, indicated as $G^\prime=f_D(H)$, where $G^\prime$ can be either reconstructed adjacency matrix, attributes or node positions. The learned embeddings $H$ can be utilized in various downstream graph mining tasks, such as node classification and node clustering.

\end{problem}

% $$$$$$$$$$$$$$$$$$$$$$$$$$$$$$$$$$$$$$$$$$$$$$$$$$$$$$$$$$
\section{Heterogeneous Graph Masked Autoencoders}

In this section, we formally present HGMAE to resolve the challenges described in \nameref{sec:intro}. In particular, HGMAE introduces two masking techniques and contains three training strategies: (1) metapath-based edge reconstruction; (2) target attribute restoration; (3) positional feature prediction. Figure \ref{fig:pipeline} illustrates the framework of HGMAE.

% ------------------------------
\subsection{Metapath-based Edge Reconstruction}

In order to capture the semantics involved in the complex graph structure, we design the metapath-based edge reconstruction strategy to explore the high-order relations through metapaths and encode the complex graph structural information. Specifically, masking the metapath-based edges breaks the short-range semantic connections between nodes, forcing the model to look elsewhere to predict the relations that are masked out. As a result, it can exploit the structure-dependency patterns more effectively and capture the high-order proximity naturally. In particular, given a heterogeneous graph $G = (\gV, \gA, T_\gV, T_\gE, X, \Phi)$, we create the metapath-based adjacency matrix $\gA^{\phi}$ for each metapath $\phi \in \Phi$ via metapath sampling \cite{mp2vec}. Since different metapaths contain different semantic information, we take them into account separately by masking and reconstructing each metapath-based adjacency matrix individually. Concretely, for each $\gA^{\phi}$, we create a binary mask following the Bernoulli distribution $M_\gA^\phi \sim \textrm{Bernoulli}(p_e)$, where $p_e < 1$ is the edge masking rate. Then, we leverage the $M_\gA^\phi$ to obtain the masked metapath-based adjacency matrix $\widetilde{\gA}^{\phi}=M_\gA^\phi \cdot \gA^{\phi}$. After that, we feed $\widetilde{\gA}^{\phi}$ and node attributes $X$ into encoder $f_E$ to obtain the latent node embeddings $H_1^{\phi}$. The process is formulated as follows:
\begin{equation}
    \begin{split}
        H_1^{\phi} = f_E(\widetilde{\gA}^{\phi}, X).
    \end{split}
\end{equation}
After that, we send $H_1^{\phi}$ with $\widetilde{\gA}^{\phi}$ into the decoder $f_D$ to generate the decoded node embeddings $H_2^{\phi}$. Later, we leverage $H_2^{\phi}$ to reconstruct the adjacency matrix: 
\begin{equation}
    H_2^{\phi} = f_D(\widetilde{\gA}^{\phi}, H_1^{\phi}), 
    \quad
    \gA^{\phi\prime} = \sigma ((H_2^{\phi})^T \cdot H_2^{\phi}), 
\end{equation}
where $\gA^{\phi\prime}$ is the reconstructed metapath-based adjacency matrix and $\sigma$ is the sigmoid activation function. Next, we compare the target $\gA^{\phi}$ and the reconstruction $\gA^{\phi\prime}$ by each node using a scaled cosine error:
\begin{equation}
    \mathcal{L}^{\phi} = \frac{1}{|\gA^{\phi}|} \sum_{v \in \gV} (1 - \frac{\gA^{\phi}_v \cdot \gA_v^{\phi\prime}}{\| \gA_v^{\phi} \| \times \| \gA_v^{\phi\prime}\|})^{\gamma_1},
\end{equation}
where $\mathcal{L}^{\phi}$ is the loss for metapath $\phi$ and $\gamma_1$ is the scaling factor. To combine $\mathcal{L}^{\phi}$ for each metapath $\phi \in \Phi$, we first introduce a semantic-level attention vector $q$ to automatically learn the important score $s^{\phi}$ of each metapath. Then, we normalize the scores of all metapaths using softmax function to obtain the weight of metapath $\phi$, denoted as $\alpha^{\phi}$. The process is shown as follows:
\begin{equation}
    s^{\phi} = q^\text{T} \cdot \tanh (W \cdot H_1^{\phi} + b),
    \quad
    \alpha^{\phi} = \frac{\exp(s^{\phi})}{\sum_{\phi \in \Phi }^{}\exp(s^{\phi})},
\end{equation}
where $W$ is the weight matrix and $b$ is the bias vector. After that, we fuse the loss for each metapath to obtain the combined loss $\mathcal{L}_{\textrm{MER}}$ for metapath-based edge reconstruction:
\begin{equation}
    \mathcal{L}_{\textrm{MER}} = \sum_{\phi \in \Phi} \alpha^{\phi} \cdot \mathcal{L}^{\phi}.
    \label{eq:loss_mer}
\end{equation}

% ------------------------------
\subsection{Target Attribute Restoration}

In order to explore the content information involved in node attributes and facilitate the model to focus on the target node type, we design the target attribute restoration strategy. In particular, we mask the attributes for nodes of target type, and let the model reconstruct the masked attributes. Considering the node attributes play a vital role in deriving the node embeddings, the model can benefit from restoring the attributes and preserving the meaningful content information in the training loop. In addition, we develop an adaptive attribute masking technique that includes dynamic mask rate, leaving unchanged and replacing to ensure more effective and stable learning.

\noindent
\textbf{Dynamic Mask Rate.}
Masking a variable percentage of attributes instead of using a fixed mask rate enables the model to maintain stable performance despite the changeable amount of input information \cite{changeable_mask_rate}. Concretely, we gradually increase the attribute mask rate according to a linear mask scheduling function so that the model can adaptively learn from easy to difficult. Formally, let $\delta(m)$ represent the mask scheduling function that computes the attribute mask rate $p_a$ with respect to training epoch $m$, where $m \in \{0,1,..., \gM\}$ and $\gM$ is the total number of training epochs. We choose $\delta(m)$ such that it is linearly increasing by step $\Delta$ for each epoch $m$, namely, $\delta(m+1)=\delta(m)+\Delta$. We further set $\delta(0)=\textrm{MIN}_{p_a}$, $\delta(\gM)=\textrm{MAX}_{p_a}$, and $\delta(m) \leq \delta(\gM)$ to ensure our model converges, where $\Delta$, $\textrm{MIN}_{p_a}$, $\textrm{MAX}_{p_a}$ are hyper-parameters adjustable over different datasets. After we get the attribute mask rate $p_a=\delta(m)$ given epoch $m$, we first sample a subset of nodes $\widetilde{\gV} \in \gV_t$ with rate $p_a$ for target node type $t$. We then mask each of their attributes using a learnable mask token $[M]$. Correspondingly, for each $v \in \gV_t$, the node attribute $\widetilde{\vx}_{v}$ in the masked attribute matrix $\widetilde{X}$ can be defined as:
\begin{equation}
    \widetilde{\vx}_{v}=
    \begin{cases}
    x_{[M]} \quad & \text{if $v\in\widetilde{\gV}$} \\
    x_v \quad & \text{if $v\notin\widetilde{\gV}$}. 
    \end{cases}
\end{equation}

\noindent
\textbf{Leaving Unchanged and Replacing.}
Due to the fact that the mask token $[M]$ does not appear during inference, a mismatch between training and inference can occur, jeopardizing the model's ability to learn \cite{mismatch_mask_token}. Therefore, we introduce the Leaving Unchanged and Replacing methods. Specifically, we first replace a percentage of mask tokens by random tokens, with the replace rate $p_{r}$. In addition, we select another percentage of nodes with rate $p_{u}$ and leave them unchanged by utilizing the origin attribute $x_v$, while $p_{u}$ indicates the leave unchanged rate. The idea of leaving unchanged and replacing is simple, yet as we will see, effective and practical, which is also the opposite of what GraphMAE claims. After that, we send node attributes $\widetilde{X}$ and graph adjacency matrix $\gA$ into encoder $f_E$ to obtain the latent node embeddings $H_3$. To enforce the encoder learns informative embeddings without relying on the decoder's capability for restoration, we apply another mask token $[DM]$ to $H_3$ before sending it into the decoder. The process is formulated as follows:
\begin{equation}
    \begin{split}
        H_3 = f_E(\gA, \widetilde{X}),
        \quad
        \widetilde{H}_3=
        \begin{cases}
        h_{[DM]} \quad & \text{if $v_i\in\widetilde{\gV}$} \\
        h_i \quad & \text{if $v_i\notin\widetilde{\gV}$}, 
        \end{cases}
    \end{split}
\end{equation}
where $\widetilde{H}_3$ denotes the masked latent node embeddings. Next, we send $\gA$ and $\widetilde{H}_3$ into the decoder $f_D$ to obtain the restored node attributes $Z$:
\begin{equation}
    \begin{split}
        Z = f_D(\gA, \widetilde{H}_3).
    \end{split}
\end{equation}
Subsequently, we define the loss of target attribute restoration $\mathcal{L}_{\textrm{TAR}}$ by comparing masked attribute matrix $\widetilde{X}$ and $Z$ with scaling factor $\gamma_2$. The loss function is described as follows:
\begin{equation}
    \mathcal{L}_{\textrm{TAR}} = \frac{1}{|\widetilde{\gV}|} \sum_{v \in \widetilde{\gV}} (1 - \frac{\widetilde{X}_v \cdot Z_v}{\| \widetilde{X}_v \| \times \| Z_v\|})^{\gamma_2}.
    \label{eq:loss_tar}
\end{equation}

% ------------- classification table -----------------
\begin{table*}[t]
  \caption{Node classification performance comparison. The best results are highlighted in bold.}
  \label{tab:classification}
  \begin{center}
  \resizebox{\textwidth}{!}{
  \begin{NiceTabular}{c|c|c|cccccccccc|c}
    \toprule
    Datasets & Metric & Split & GraphSAGE & GAE & Mp2vec & HERec & HetGNN & HAN & DGI & DMGI & HeCo & GraphMAE & HGMAE\\
    \midrule
    
    % ---------------------------
    \multirow{10}{*}{DBLP}
    &\multirow{3}{*}{Mi-F1}
    &20&71.44$\pm$8.7&91.55$\pm$0.1&89.67$\pm$0.1&90.24$\pm$0.4&90.11$\pm$1.0&90.16$\pm$0.9&88.72$\pm$2.6&90.78$\pm$0.3&91.97$\pm$0.2 & 89.31$\pm$0.7 & \textbf{92.71$\pm$0.5} \\
    &&40&73.61$\pm$8.6&90.00$\pm$0.3&89.14$\pm$0.2&90.15$\pm$0.4&89.03$\pm$0.7&89.47$\pm$0.9&89.22$\pm$0.5&89.92$\pm$0.4&90.76$\pm$0.3 & 87.80$\pm$0.5 & \textbf{92.43$\pm$0.3} \\
    &&60&74.05$\pm$8.3&90.95$\pm$0.2&91.17$\pm$0.1&91.01$\pm$0.3&90.43$\pm$0.6&90.34$\pm$0.8&90.35$\pm$0.8&90.66$\pm$0.5&91.59$\pm$0.2 & 89.82$\pm$0.4 & \textbf{93.05$\pm$0.3} \\
    \cmidrule{2-14}
    &\multirow{3}{*}{Ma-F1}
    &20&71.97$\pm$8.4&90.90$\pm$0.1&88.98$\pm$0.2&89.57$\pm$0.4&89.51$\pm$1.1&89.31$\pm$0.9&87.93$\pm$2.4&89.94$\pm$0.4&91.28$\pm$0.2 & 87.94$\pm$0.7 & \textbf{92.28$\pm$0.5} \\
    &&40&73.69$\pm$8.4&89.60$\pm$0.3&88.68$\pm$0.2&89.73$\pm$0.4&88.61$\pm$0.8&88.87$\pm$1.0&88.62$\pm$0.6&89.25$\pm$0.4&90.34$\pm$0.3 & 86.85$\pm$0.7 & \textbf{92.12$\pm$0.3} \\
    &&60&73.86$\pm$8.1&90.08$\pm$0.2&90.25$\pm$0.1&90.18$\pm$0.3&89.56$\pm$0.5&89.20$\pm$0.8&89.19$\pm$0.9&89.46$\pm$0.6&90.64$\pm$0.3 & 88.07$\pm$0.6 & \textbf{92.33$\pm$0.3} \\
    \cmidrule{2-14}
    &\multirow{3}{*}{AUC}
    &20&90.59$\pm$4.3&98.15$\pm$0.1&97.69$\pm$0.0&98.21$\pm$0.2&97.96$\pm$0.4&98.07$\pm$0.6&96.99$\pm$1.4&97.75$\pm$0.3&98.32$\pm$0.1 & 92.23$\pm$3.0 & \textbf{98.90$\pm$0.1} \\
    &&40&91.42$\pm$4.0&97.85$\pm$0.1&97.08$\pm$0.0&97.93$\pm$0.1&97.70$\pm$0.3&97.48$\pm$0.6&97.12$\pm$0.4&97.23$\pm$0.2&98.06$\pm$0.1 & 91.76$\pm$2.5 & \textbf{98.55$\pm$0.1} \\
    &&60&91.73$\pm$3.8&98.37$\pm$0.1&98.00$\pm$0.0&98.49$\pm$0.1&97.97$\pm$0.2&97.96$\pm$0.5&97.76$\pm$0.5&97.72$\pm$0.4&98.59$\pm$0.1 & 91.63$\pm$2.5 & \textbf{98.89$\pm$0.1} \\
    \midrule
    
    % ---------------------------
    \multirow{10}{*}{Freebase}
    &\multirow{3}{*}{Mi-F1}
    &20&54.83$\pm$3.0&55.20$\pm$0.7&56.23$\pm$0.8&57.92$\pm$0.5&56.85$\pm$0.9&57.24$\pm$3.2&58.16$\pm$0.9&58.26$\pm$0.9&61.72$\pm$0.6 & 64.88$\pm$1.8 & \textbf{65.15$\pm$1.3} \\
    &&40&57.08$\pm$3.2&56.05$\pm$2.0&61.01$\pm$1.3&62.71$\pm$0.7&53.96$\pm$1.1&63.74$\pm$2.7&57.82$\pm$0.8&54.28$\pm$1.6&64.03$\pm$0.7 & 62.34$\pm$1.0 & \textbf{67.23$\pm$0.8} \\
    &&60&55.92$\pm$3.2&53.85$\pm$0.4&58.74$\pm$0.8&58.57$\pm$0.5&56.84$\pm$0.7&61.06$\pm$2.0&57.96$\pm$0.7&56.69$\pm$1.2&63.61$\pm$1.6 & 59.48$\pm$6.2 & \textbf{67.44$\pm$1.2} \\
    \cmidrule{2-14}
    &\multirow{3}{*}{Ma-F1}
    &20&45.14$\pm$4.5&53.81$\pm$0.6&53.96$\pm$0.7&55.78$\pm$0.5&52.72$\pm$1.0&53.16$\pm$2.8&54.90$\pm$0.7&55.79$\pm$0.9&59.23$\pm$0.7 & 59.04$\pm$1.0 & \textbf{62.06$\pm$1.0} \\
    &&40&44.88$\pm$4.1&52.44$\pm$2.3&57.80$\pm$1.1&59.28$\pm$0.6&48.57$\pm$0.5&59.63$\pm$2.3&53.40$\pm$1.4&49.88$\pm$1.9&61.19$\pm$0.6 & 56.40$\pm$1.1 & \textbf{64.64$\pm$0.9} \\
    &&60&45.16$\pm$3.1&50.65$\pm$0.4&55.94$\pm$0.7&56.50$\pm$0.4&52.37$\pm$0.8&56.77$\pm$1.7&53.81$\pm$1.1&52.10$\pm$0.7&60.13$\pm$1.3 & 51.73$\pm$2.3 & \textbf{63.84$\pm$1.0} \\
    \cmidrule{2-14}
    &\multirow{3}{*}{AUC}
    &20&67.63$\pm$5.0&73.03$\pm$0.7&71.78$\pm$0.7&73.89$\pm$0.4&70.84$\pm$0.7&73.26$\pm$2.1&72.80$\pm$0.6&73.19$\pm$1.2&76.22$\pm$0.8 & 72.60$\pm$0.2 & \textbf{78.36$\pm$1.1} \\
    &&40&66.42$\pm$4.7&74.05$\pm$0.9&75.51$\pm$0.8&76.08$\pm$0.4&69.48$\pm$0.2&77.74$\pm$1.2&72.97$\pm$1.1&70.77$\pm$1.6&78.44$\pm$0.5 & 72.44$\pm$1.6 & \textbf{79.69$\pm$0.7} \\
    &&60&66.78$\pm$3.5&71.75$\pm$0.4&74.78$\pm$0.4&74.89$\pm$0.4&71.01$\pm$0.5&75.69$\pm$1.5&73.32$\pm$0.9&73.17$\pm$1.4&78.04$\pm$0.4 & 70.66$\pm$1.6 & \textbf{79.11$\pm$1.3} \\
    \midrule
    
    % ---------------------------
    \multirow{10}{*}{ACM}
    &\multirow{3}{*}{Mi-F1}
    &20&49.72$\pm$5.5&68.02$\pm$1.9&53.13$\pm$0.9&57.47$\pm$1.5&71.89$\pm$1.1&85.11$\pm$2.2&79.63$\pm$3.5&87.60$\pm$0.8&88.13$\pm$0.8 & 82.48$\pm$1.9 & \textbf{90.24$\pm$0.5} \\
    &&40&60.98$\pm$3.5&66.38$\pm$1.9&64.43$\pm$0.6&62.62$\pm$0.9&74.46$\pm$0.8&87.21$\pm$1.2&80.41$\pm$3.0&86.02$\pm$0.9&87.45$\pm$0.5 & 82.93$\pm$1.1 & \textbf{90.18$\pm$0.6} \\
    &&60&60.72$\pm$4.3&65.71$\pm$2.2&62.72$\pm$0.3&65.15$\pm$0.9&76.08$\pm$0.7&88.10$\pm$1.2&80.15$\pm$3.2&87.82$\pm$0.5&88.71$\pm$0.5 & 80.77$\pm$1.1 & \textbf{91.34$\pm$0.4} \\
    \cmidrule{2-14}
    &\multirow{3}{*}{Ma-F1}
    &20&47.13$\pm$4.7&62.72$\pm$3.1&51.91$\pm$0.9&55.13$\pm$1.5&72.11$\pm$0.9&85.66$\pm$2.1&79.27$\pm$3.8&87.86$\pm$0.2&88.56$\pm$0.8 & 82.26$\pm$1.5 & \textbf{90.66$\pm$0.4} \\
    &&40&55.96$\pm$6.8&61.61$\pm$3.2&62.41$\pm$0.6&61.21$\pm$0.8&72.02$\pm$0.4&87.47$\pm$1.1&80.23$\pm$3.3&86.23$\pm$0.8&87.61$\pm$0.5 & 82.00$\pm$1.1 & \textbf{90.15$\pm$0.6} \\
    &&60&56.59$\pm$5.7&61.67$\pm$2.9&61.13$\pm$0.4&64.35$\pm$0.8&74.33$\pm$0.6&88.41$\pm$1.1&80.03$\pm$3.3&87.97$\pm$0.4&89.04$\pm$0.5 & 80.29$\pm$1.0 & \textbf{91.59$\pm$0.4} \\
    \cmidrule{2-14}
    &\multirow{3}{*}{AUC}
    &20&65.88$\pm$3.7&79.50$\pm$2.4&71.66$\pm$0.7&75.44$\pm$1.3&84.36$\pm$1.0&93.47$\pm$1.5&91.47$\pm$2.3&96.72$\pm$0.3&96.49$\pm$0.3 & 92.09$\pm$0.5 & \textbf{97.69$\pm$0.1} \\
    &&40&71.06$\pm$5.2&79.14$\pm$2.5&80.48$\pm$0.4&79.84$\pm$0.5&85.01$\pm$0.6&94.84$\pm$0.9&91.52$\pm$2.3&96.35$\pm$0.3&96.40$\pm$0.4 & 92.65$\pm$0.5 & \textbf{97.52$\pm$0.1} \\
    &&60&70.45$\pm$6.2&77.90$\pm$2.8&79.33$\pm$0.4&81.64$\pm$0.7&87.64$\pm$0.7&94.68$\pm$1.4&91.41$\pm$1.9&96.79$\pm$0.2&96.55$\pm$0.3 & 91.49$\pm$0.6 & \textbf{97.87$\pm$0.1} \\
    \midrule
    
    % ---------------------------
    \multirow{10}{*}{AMiner}
    &\multirow{3}{*}{Mi-F1}
    &20&49.68$\pm$3.1&65.78$\pm$2.9&60.82$\pm$0.4&63.64$\pm$1.1&61.49$\pm$2.5&68.86$\pm$4.6&62.39$\pm$3.9&63.93$\pm$3.3&78.81$\pm$1.3 & 68.21$\pm$0.3 & \textbf{80.30$\pm$0.7} \\
    &&40&52.10$\pm$2.2&71.34$\pm$1.8&69.66$\pm$0.6&71.57$\pm$0.7&68.47$\pm$2.2&76.89$\pm$1.6&63.87$\pm$2.9&63.60$\pm$2.5&80.53$\pm$0.7 & 74.23$\pm$0.2 & \textbf{82.35$\pm$1.0} \\
    &&60&51.36$\pm$2.2&67.70$\pm$1.9&63.92$\pm$0.5&69.76$\pm$0.8&65.61$\pm$2.2&74.73$\pm$1.4&63.10$\pm$3.0&62.51$\pm$2.6&\textbf{82.46$\pm$1.4} & 72.28$\pm$0.2 & 81.69$\pm$0.6 \\
    \cmidrule{2-14}
    &\multirow{3}{*}{Ma-F1}
    &20&42.46$\pm$2.5&60.22$\pm$2.0&54.78$\pm$0.5&58.32$\pm$1.1&50.06$\pm$0.9&56.07$\pm$3.2&51.61$\pm$3.2&59.50$\pm$2.1&71.38$\pm$1.1 & 62.64$\pm$0.2 & \textbf{72.28$\pm$0.6} \\
    &&40&45.77$\pm$1.5&65.66$\pm$1.5&64.77$\pm$0.5&64.50$\pm$0.7&58.97$\pm$0.9&63.85$\pm$1.5&54.72$\pm$2.6&61.92$\pm$2.1&73.75$\pm$0.5 & 68.17$\pm$0.2 & \textbf{75.27$\pm$1.0} \\
    &&60&44.91$\pm$2.0&63.74$\pm$1.6&60.65$\pm$0.3&65.53$\pm$0.7&57.34$\pm$1.4&62.02$\pm$1.2&55.45$\pm$2.4&61.15$\pm$2.5&\textbf{75.80$\pm$1.8} & 68.21$\pm$0.2 & 74.67$\pm$0.6 \\
    \cmidrule{2-14}
    &\multirow{3}{*}{AUC}
    &20&70.86$\pm$2.5&85.39$\pm$1.0&81.22$\pm$0.3&83.35$\pm$0.5&77.96$\pm$1.4&78.92$\pm$2.3&75.89$\pm$2.2&85.34$\pm$0.9&90.82$\pm$0.6 & 86.29$\pm$4.1 & \textbf{93.22$\pm$0.6} \\
    &&40&74.44$\pm$1.3&88.29$\pm$1.0&88.82$\pm$0.2&88.70$\pm$0.4&83.14$\pm$1.6&80.72$\pm$2.1&77.86$\pm$2.1&88.02$\pm$1.3&92.11$\pm$0.6 & 89.98$\pm$0.0 & \textbf{94.68$\pm$0.4} \\
    &&60&74.16$\pm$1.3&86.92$\pm$0.8&85.57$\pm$0.2&87.74$\pm$0.5&84.77$\pm$0.9&80.39$\pm$1.5&77.21$\pm$1.4&86.20$\pm$1.7&92.40$\pm$0.7 & 88.32$\pm$0.0 & \textbf{94.59$\pm$0.3} \\
    \bottomrule
  \end{NiceTabular}}
  \end{center}
\end{table*}

% ------------------------------
\subsection{Positional Feature Prediction}

In order to incorporate the positional information for each node, we design the positional feature prediction strategy. In particular, we first extract metapath aware positional features $P$ using Mp2vec \cite{mp2vec}. Then, we let the model predict the positional features while only using the latent node embeddings $H_3$ from the encoder $f_E$. To encourage the encoder to take positional information into account during training, we abandon the graph structure input for decoding and instead employ the traditional multilayer perceptron (MLP) $f_{\textrm{MLP}}$ as the decoder. Correspondingly, the encoder is able to capture node positional information within a broader context of the graph structure. The process is formulated as follows:
\begin{equation}
    \begin{split}
        P^{\prime} = f_{\textrm{MLP}}(H_3),
    \end{split}
\end{equation}
where $P^{\prime}$ denotes the predicted node positional features. Then, the loss of positional feature prediction $\mathcal{L}_{\textrm{PFP}}$ is determined by comparing the target positional features $P$ and the predicted positional features $P^{\prime}$:
\begin{equation}
    \mathcal{L}_{\textrm{PFP}} = \frac{1}{|\gV|} \sum_{v \in \gV} (1 - \frac{P_v \cdot P_v^{\prime}}{\| P_v \| \times \| P_v^{\prime}\|})^{\gamma_3},
    \label{eq:loss_pfp}
\end{equation}
where $\gamma_3$ is the scaling factor. The final objective function $\mathcal{L}$ is defined as the weighted combination of the metapath-based edge reconstruction loss $\mathcal{L}_{\textrm{MER}}$, the target attribute restoration loss $\mathcal{L}_{\textrm{TAR}}$ and the positional feature prediction loss $\mathcal{L}_{\textrm{PFP}}$:
\begin{equation}
    \mathcal{L} = \lambda \mathcal{L}_{\textrm{MER}} + \mu \mathcal{L}_{\textrm{TAR}} + \eta \mathcal{L}_{\textrm{PFP}},
    \label{eq:final_loss}
\end{equation}
where $\lambda$, $\mu$ and $\eta$ are trade-off weights for balancing $\mathcal{L}_{\textrm{MER}}$, $\mathcal{L}_{\textrm{TAR}}$ and $\mathcal{L}_{\textrm{PFP}}$, respectively.

% $$$$$$$$$$$$$$$$$$$$$$$$$$$$$$$$$$$$$$$$$$$$$$$$$$$$$$$$$$
\section{Experiments}

In this section, we conduct extensive experiments to compare the performances of different models. We also show ablation studies, parameter sensitivity, and embedding visualization to demonstrate the superiority of HGMAE.

% ------------------------------
\subsection{Experimental Setup}

\textbf{Datasets and Baselines.}
We employ four real datasets to evaluate the proposed model, including DBLP \cite{magnn}, Freebase \cite{freebase}, ACM \cite{acm}, and AMiner \cite{aminer}. We compare with 10 baselines including unsupervised homogeneous methods GraphSAGE \cite{graphsage}, GAE \cite{gae}, DGI \cite{dgi}, GraphMAE \cite{graphmae}, unsupervised/semi-supervised heterogeneous methods Mp2vec \cite{mp2vec}, HERec \cite{herec}, HetGNN \cite{hetgnn}, DMGI \cite{dmgi}, HeCo \cite{heco}, and HAN \cite{han}.

\noindent
\textbf{Implementation Details.}
For baselines, we adhere to the settings described in their original papers and follow the setup in HeCo \cite{heco}. For the proposed HGMAE, we use HAN \cite{han} as the default encoder and decoder. We search the learning rate from 1e-4 to 5e-3, tune the patience for early stopping from 5 to 20, and test the leave unchanged and replaced rates from 0 to 0.5 with step 0.1. For dynamic mask rate, we set $\textrm{MIN}_{p_a}$ to 0.5, $\textrm{MAX}_{p_a}$ to 0.8 and $\Delta$ equals 0.005. For all methods, we report the mean and standard deviation of 10 runs with different random seeds.

% ------------------------------
\subsection{Node Classification}

We first evaluate different models for the node classification task and report their performances in Table \ref{tab:classification}. Specifically, we use 20, 40, 60 labeled nodes per class as training set and 1000 nodes each for validation and test sets. We use Micro-F1, Macro-F1 and AUC as evaluation metrics. According to the table, we find that our model HGMAE outperforms all the baselines across various datasets except for few cases. In AMiner, HGMAE has a small difference compared to the best baseline HeCo on few F1 values, but still remains the second best among all models. Considering AMiner has fewer edges and metapaths compared to other datasets, it is understandable that HGMAE could be biased with simple generative losses, while HeCo performs better by employing the complex contrastive loss on intricate graph views. However, HGMAE is more stable and performs better across various datasets and experimental settings. In addition, GraphMAE performs poorly on all datasets. This further demonstrates the significance of addressing the identified challenges and proves the effectiveness of our model.

% ------------- clustering table -----------------
\begin{table}
  \caption{Node clustering performance comparison. The best results are highlighted in bold.}
  \label{tab:clustering}
  \begin{center}
  \resizebox{0.47\textwidth}{!}{
  \begin{NiceTabular}{c|cc|cc|cc|cc}
    \toprule
    Datasets & \multicolumn{2}{c|}{DBLP} & \multicolumn{2}{c|}{Freebase} & \multicolumn{2}{c|}{ACM} & \multicolumn{2}{c}{AMiner} \\
    \midrule
    Metrics & NMI & ARI & NMI & ARI & NMI & ARI & NMI & ARI\\
    \midrule
    GraphSage & 51.50 & 36.40 & 9.05 & 10.49 & 29.20 & 27.72 & 15.74 & 10.10\\
    GAE & 72.59 & 77.31 & 19.03 & 14.10 & 27.42 & 24.49 & 28.58 & 20.90\\
    Mp2vec & 73.55 & 77.70 & 16.47 & 17.32 & 48.43 & 34.65 & 30.80 & 25.26\\
    HERec & 70.21 & 73.99 & 19.76 & 19.36 & 47.54 & 35.67 & 27.82 & 20.16\\
    HetGNN & 69.79 & 75.34 & 12.25 & 15.01 & 41.53 & 34.81 & 21.46 & 26.60\\
    DGI & 59.23 & 61.85 & 18.34 & 11.29 & 51.73 & 41.16 & 22.06 & 15.93\\
    DMGI & 70.06 & 75.46 & 16.98 & 16.91 & 51.66 & 46.64 & 19.24 & 20.09\\
    HeCo & 74.51 & 80.17 & 20.38 & 20.98 & 56.87 & 56.94 & 32.26 & 28.64\\
    GraphMAE & 65.86 & 69.75 & 19.43 & 20.05 & 47.03 & 46.48 & 17.98 & 21.52\\
    \midrule
    HGMAE & \textbf{76.92} & \textbf{82.34} & \textbf{22.05} & \textbf{22.84} & \textbf{66.68} & \textbf{71.51} & \textbf{41.10} & \textbf{38.27} \\
    \bottomrule
  \end{NiceTabular}}
  \end{center}
\end{table}

% ------------------------------
\subsection{Node Clustering}

We further evaluate different models for node clustering task and report their performances in Table \ref{tab:clustering}. In particular, we apply K-means as the learning algorithm and utilize normalized mutual information (NMI) and adjusted rand index (ARI) as the evaluation metrics. From the table, we can find that HGMAE consistently achieves the best results on all datasets, which validates the effectiveness of HGMAE from a different perspective. Concretely, HGMAE maintains superior learning capability and outperforms existing baselines by a large margin in most cases. For example, compared with the best baseline HeCo, HGMAE improves NMI by +17\% and ARI by +25\% on ACM dataset, as well as 27\% and 33\% on AMiner dataset. This further demonstrates the superiority of our model.

% ------------------------------
\subsection{Ablation Study}

Since HGMAE contains various training strategies (i.e., metapath-based edge reconstruction (MER), target attribute restoration (TAR), and positional feature prediction (PFP)), we conduct ablation studies on the node classification task to analyze the contributions of different strategies by removing each of them independently (see Table \ref{tab:ablation}). Specifically, removing MER significantly affects the performance, showing that MER has large contribution to HGMAE. In addition, the decreasing performances of removing TAR and PFP demonstrate the effectiveness of TAR and PFP in enhancing the model, respectively. Finally, HGMAE achieves the best results in all cases, indicating the strong capability of different strategies in our model.

% ------------- ablation table -----------------
\begin{table}
  \caption{Results of different model variants.}
  \label{tab:ablation}
  \begin{center}
  \resizebox{0.44\textwidth}{!}{
  \begin{NiceTabular}{cc|c|c|c|c}
    \toprule
    Datasets & Metric & w/o MER & w/o TAR & w/o PFP & HGMAE \\
    \midrule
    
    \multirow{3}{*}{DBLP} 
     & Mi-F1 & 91.99$\pm$0.5 & 92.06$\pm$0.5 & 92.21$\pm$0.6 & \textbf{92.73$\pm$0.4} \\
     & Ma-F1 & 91.39$\pm$0.5 & 91.53$\pm$0.6 & 91.68$\pm$0.5 & \textbf{92.24$\pm$0.4} \\
     & AUC & 98.60$\pm$0.1 & 98.66$\pm$0.1 & 98.66$\pm$0.1 & \textbf{98.78$\pm$0.1} \\
    \midrule
    
    \multirow{3}{*}{Freebase} 
     & Mi-F1 & 66.01$\pm$0.8 & 64.76$\pm$1.6 & 66.20$\pm$1.6 & \textbf{66.61$\pm$1.1} \\
     & Ma-F1 & 63.16$\pm$0.7 & 61.12$\pm$1.1 & 62.10$\pm$1.3 & \textbf{63.51$\pm$1.0} \\ 
     & AUC & 78.04$\pm$0.6 & 78.71$\pm$1.0 & 77.92$\pm$0.9 & \textbf{79.05$\pm$1.0} \\
    \midrule
    
    \multirow{3}{*}{ACM} 
     & Mi-F1 & 76.85$\pm$0.2 & 88.54$\pm$0.4 & 89.81$\pm$0.5 & \textbf{90.59$\pm$0.5} \\
     & Ma-F1 & 71.93$\pm$0.4 & 88.82$\pm$0.4 & 89.94$\pm$0.4 & \textbf{90.80$\pm$0.5} \\
     & AUC & 84.84$\pm$1.5 & 96.47$\pm$0.1 & 97.22$\pm$0.1 & \textbf{97.69$\pm$0.1} \\
    \midrule
    
    \multirow{3}{*}{AMiner} 
     & Mi-F1 & 72.12$\pm$1.6 & 81.22$\pm$0.7 & 80.88$\pm$1.0 & \textbf{81.45$\pm$0.8} \\
     & Ma-F1 & 65.01$\pm$1.1 & 73.66$\pm$0.7 & 73.66$\pm$0.9 & \textbf{74.07$\pm$0.7} \\
     & AUC & 85.59$\pm$1.6 & 93.70$\pm$0.3 & 93.65$\pm$0.3 & \textbf{94.16$\pm$0.4} \\
    \bottomrule
  \end{NiceTabular}}
  \end{center}
\end{table}

% -------------- figure: mask rate ----------------
\begin{figure}
	\centering
	\includegraphics[width=\columnwidth]{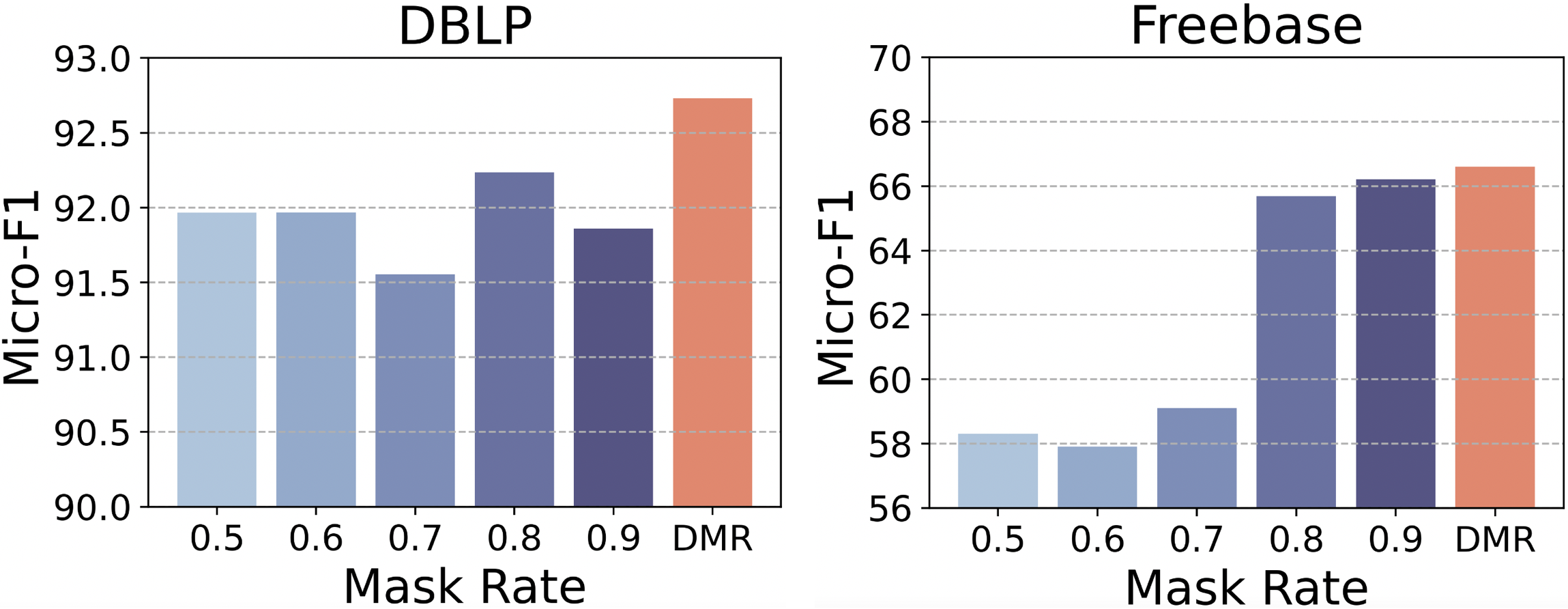}
	\caption{
	Performance of HGMAE \textit{w.r.t.} different mask rates.
	}
	\label{fig:mask_rate}
\end{figure}

% ------------------------------
\subsection{Impact of Dynamic Mask Rate}

A salient property of HGMAE is the incorporation of the dynamic mask rate (DMR), which masks a variable percentage of attributes to encourage the stable learning across various masking settings. To better understand the effectiveness of DMR, we provide a detailed analysis by comparing the performance of DMR with fixed mask rates, as shown in Figure \ref{fig:mask_rate}. We find that DMR can always achieve the best result, demonstrating its superiority compared to other mask rates. In general, the performance increases when increasing the mask rate, which indicates that a suitable mask rate will lead to better model performances. However, a mask rate that is suitable for one dataset might not be the optimal choice for other datasets. For example, without DMR, fixed mask rate 0.9 is the optimal rate for Freebase, but it performs poorly on DBLP. Consequently, it is essential that the model learns adaptively across various mask rates. In addition, DMR enables the model to learn effectively without the need for manual mask rate determination.

% ------------------------------
\subsection{Impact of Leaving Unchanged and Replacing}
\begin{figure}
	\centering
	\includegraphics[width=\columnwidth]{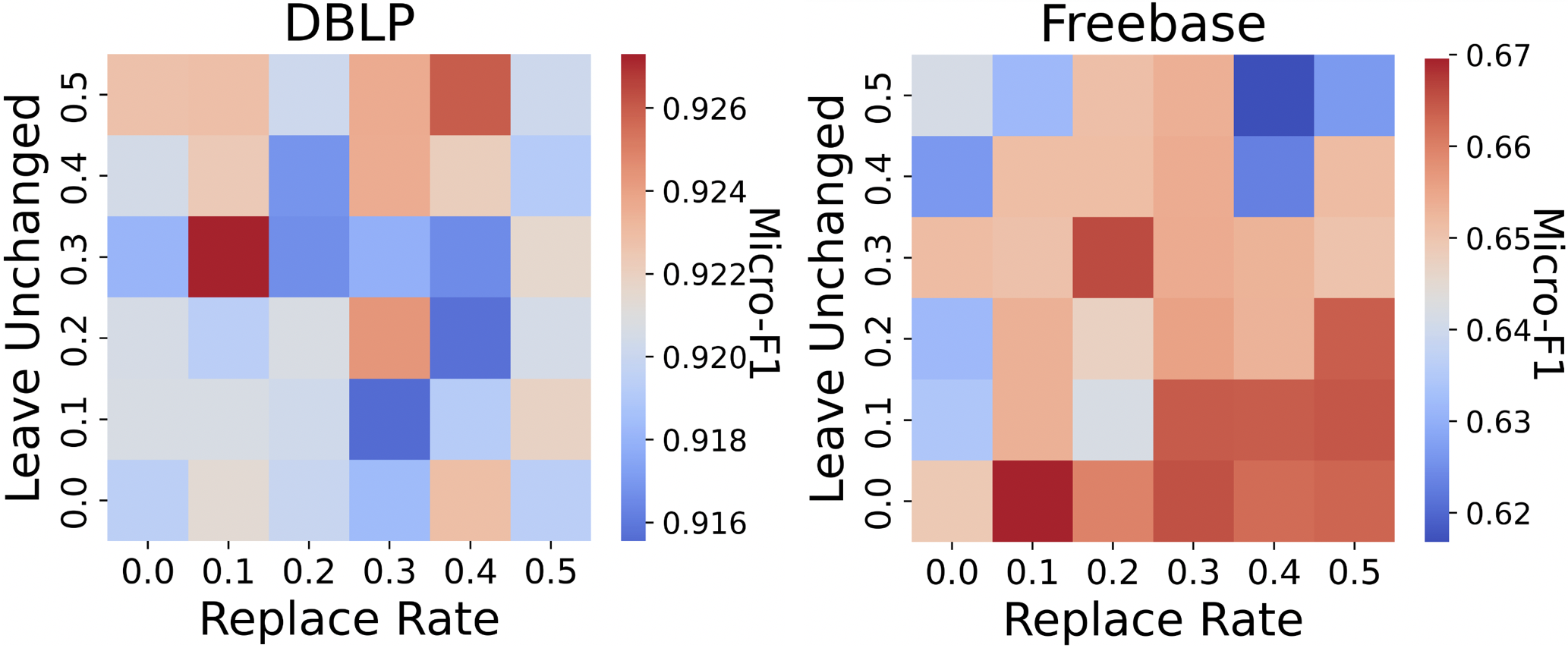}
	\caption{
	Performance of HGMAE \textit{w.r.t.} leave unchanged and replace rates.
	}
	\label{fig:heat_map}
\end{figure}

We conduct experiments to show the effectiveness of leaving unchanged and replacing in Figure \ref{fig:heat_map}. We surprisedly discover that the use of these two techniques always improves the model's ability to learn better node embeddings, which is opposite to the statement in GraphMAE \cite{graphmae}. Specifically, HGMAE achieves the optimal performance in DBLP when leave unchanged and replace rates are set to 0.3 and 0.1, and performs the best in Freebase when they are set to 0.0 and 0.1, respectively. In general, we observe that for DBLP, a high leave unchanged rate and a low replace rate are usually beneficial, whereas for Freebase, the opposite is true. This demonstrates the effectiveness of leaving unchanged and replacing, and indicates that different rates may have different impacts across datasets. Correspondingly, determining appropriate rates can result in a better model performance.

% -------------- figure: heat map ----------------

% -------------- hidden dim ----------------
\begin{figure}
	\centering
	\includegraphics[width=\columnwidth]{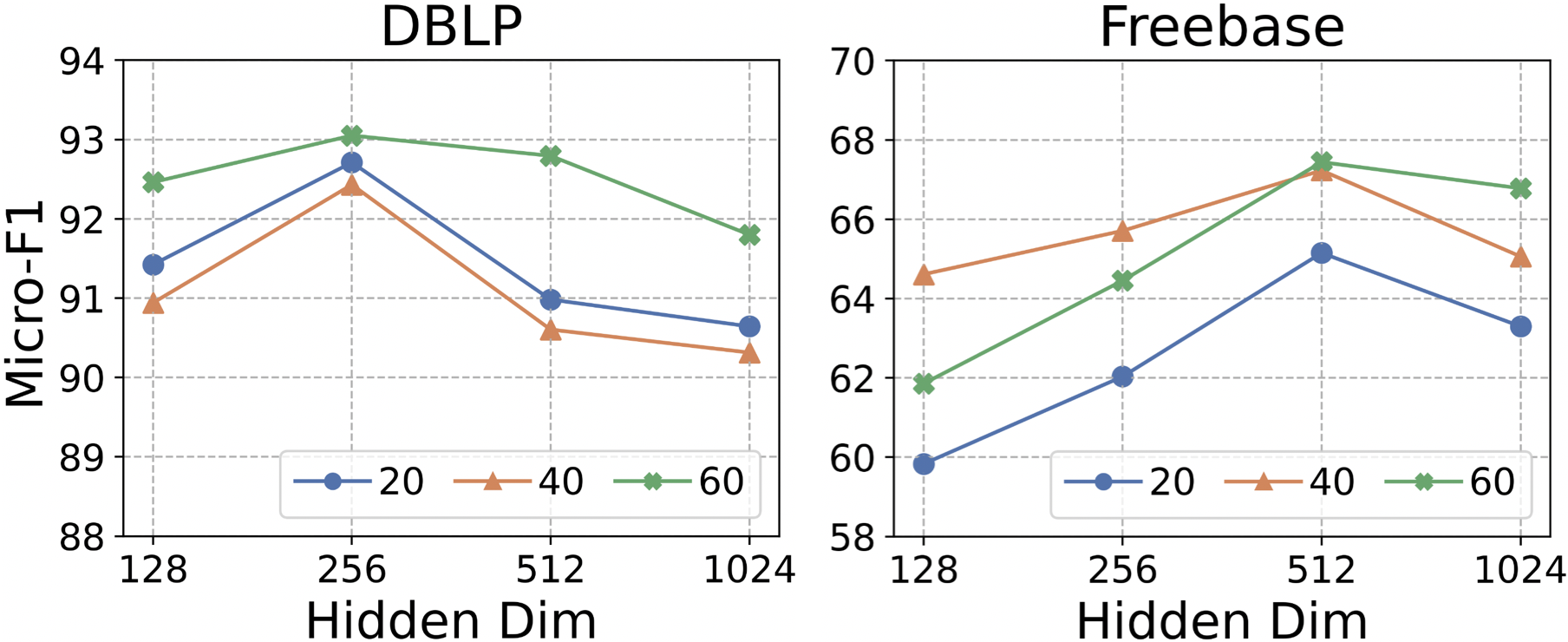}
	\caption{
	Performance of HGMAE \textit{w.r.t.} hidden dimensions.
	}
	\label{fig:hidden_dim}
\end{figure}

% -------------- figure: visualization ----------------
\begin{figure}
	\centering
	\includegraphics[width=0.4\textwidth]{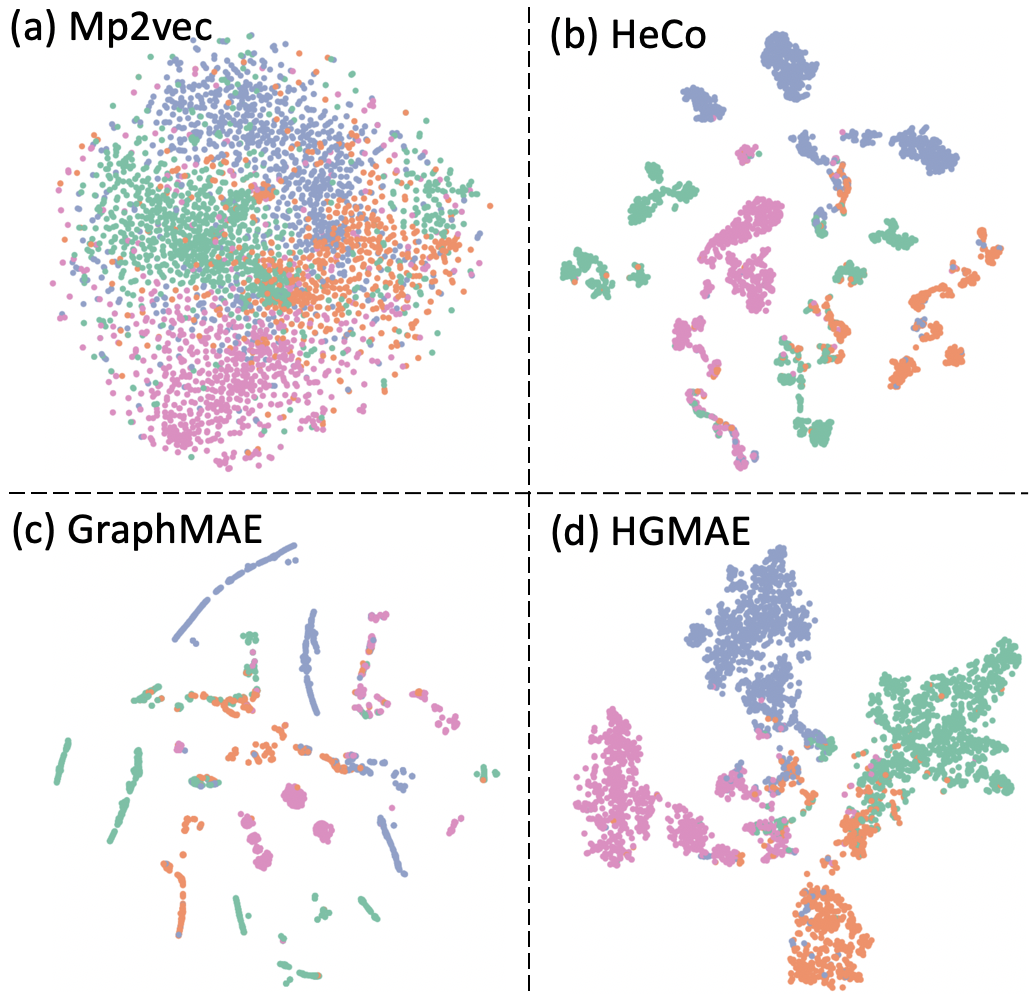}
	\caption{
	Embedding visualization of DBLP dataset. Different colors indicate different node category labels.
	}
	\label{fig:visualization}
\end{figure}

% ------------------------------
\subsection{Parameter Sensitivity}

We further perform parameters sensitivity analysis to show the impact of hidden dimensions in Figure \ref{fig:hidden_dim}. In particular, we search the number of hidden dimensions from \{128, 256, 512, 1024\}. By analyzing the figure, we find that the optimal hidden dimension can be different across datasets. In addition, increasing the hidden dimension generally enhances the performance. We ascribe this improvement to the comprehensive modeling of the data itself, while using a small hidden dimension could prevent the model from fully capturing the knowledge. However, further increasing the hidden dimension (e.g., 1024) degrades the performance. This is because applying a too wide hidden layer could decentralize the model's focus on meaningful information. Therefore, using the standard hidden dimensions (e.g., 256 or 512)  is sufficient for the proposed model to capture complex information and achieve superior performance.

% ------------------------------
\subsection{Embedding Visualization}

For a more intuitive understanding and comparison, we visualize the learned node embeddings of different models using t-SNE. As shown in Figure \ref{fig:visualization}, Mp2vec does not perform well. Nodes from different categories are mixed together. HeCo can  successfully distinguish different categories, but fail to cluster nodes that share the same category. GraphMAE can separate each category well, but the nodes tend to stay close to each other and form small chunks, while the difference between chunks is large, even when chunks belong to the same category. However, our model HGMAE can clearly identify each category and maintain a clear boundary between them. In addition, nodes with the same category form an exclusive dense cluster, instead of splitting into numerous chunks as shown in other methods. This again demonstrates the effectiveness of HGMAE and the capability of learning discriminative node embeddings.

% $$$$$$$$$$$$$$$$$$$$$$$$$$$$$$$$$$$$$$$$$$$$$$$$$$$$$$$$$$
\section{Conclusion}

In this paper, we propose and formalize the problem of generative self-supervised learning on heterogeneous graphs. To solve this problem, we propose HGMAE, a novel heterogeneous graph masked autoencoder model. HGMAE jointly considers the complex graph structure, various node attributes, and different node positions via two innovative masking techniques and three unique training strategies. Extensive experiments on multiple datasets and several tasks demonstrate the superiority of HGMAE compared to state-of-the-art methods.

\bibliography{main}

\end{document}